\documentclass[10pt, a4paper]{article}
\usepackage{lrec-coling2024} 

\usepackage{natbib}
\usepackage{multibib}
\makeatletter
\def\@mb@citenamelist{cite,citep,citet,citealp,citealt,citepalias,citetalias}
\makeatother
\newcites{languageresource}{~}

\usepackage{graphicx}
\usepackage{tabularx}
\usepackage{soul}

\usepackage{xcolor}
\usepackage{hyperref}
 \definecolor{darkblue}{rgb}{0, 0, 0.5}
  \hypersetup{colorlinks=true, citecolor=darkblue, linkcolor=darkblue, urlcolor=darkblue}

\usepackage{xstring}

\usepackage{color}
\usepackage[inline]{enumitem}
\usepackage{hyperref}

\newcommand{\todo}[1]{\textcolor{red}{TODO: #1}}

\title{Cross-Lingual Transfer Robustness to Lower-Resource Languages on Adversarial Datasets}

\name{Shadi Manafi and Nikhil Krishnaswamy} 

\address{Colorado State University \\
         Fort Collins, CO, USA \\
         \{shadim,nkrishna\}@colostate.edu}

\abstract{
Multilingual Language Models (MLLMs) exhibit robust cross-lingual transfer capabilities, or the ability to leverage information acquired in a source language and apply it to a target language. These capabilities find practical applications in well-established Natural Language Processing (NLP) tasks such as Named Entity Recognition (NER). This study aims to investigate the effectiveness of a source language when applied to a target language, particularly in the context of perturbing the input test set. We evaluate on 13 pairs of languages, each including one high-resource language (HRL) and one low-resource language (LRL) with a geographic, genetic, or borrowing relationship. We evaluate two well-known MLLMs---MBERT and XLM-R---on these pairs, in native LRL and cross-lingual transfer settings, in two tasks, under a set of different perturbations. Our findings indicate that NER cross-lingual transfer depends largely on the overlap of entity chunks. If a source and target language have more entities in common, the transfer ability is stronger. Models using cross-lingual transfer also appear to be somewhat more robust to certain perturbations of the input, perhaps indicating an ability to leverage stronger representations derived from the HRL.  Our research provides valuable insights into cross-lingual transfer and its implications for NLP applications, and underscores the need to consider linguistic nuances and potential limitations when employing MLLMs across distinct languages.
 \\ \newline  }

\begin{document}

\maketitleabstract

\section{Introduction}

Multilingual Language Models (MLLMs) such as MBERT \cite{devlin-etal-2019-bert} and XLM-R \cite{conneau2020unsupervised}, demonstrate strong cross-lingual transfer abilities for downstream tasks \cite{hu2020xtreme}. Cross-lingual transfer leverages information from a source language, improving performance in target languages, which often leads to impressive performance on the same task in other languages. This can be particularly beneficial for NLP task performance in low-resource languages (LRLs). \citet{pires2019multilingual} first observed this phenomenon for NER and POS tagging tasks. Concurrently, \citet{wu2019beto} also demonstrated MBERT's cross-lingual transfer ability on tasks such as Document Classification, Natural Language Inference, and Dependency Parsing.

However, challenges persist in instances of common tasks like NER, even for allegedly ``generalist'' models like ChatGPT \cite{wu2023bloomberggpt, qin2023chatgpt}. Zero-shot learning may sometimes rely on ``vocabulary memorization'' rather than true language understanding \cite{patil-etal-2022-overlap}, and determining whether and why this is the case on specific tasks remains challenging due to linguistic variations and domain-specific differences.


Recent research highlights sensitivity of NLP tasks to minor input changes, in contrast to evaluating against fixed gold standard \cite{gardner2020evaluating}.

In this paper, we evaluate specifically the extent to which MLLM performance on a high-resource language (HRL) can be expected to transfer to a LRL that shares vocabulary similarity due to areal or genetic proximity, or linguistic borrowing, and the extent to which this assistance in performance is robust to input changes that are driven either by particulars of the task, or by semantic similarity.


Hence, we explore the following questions in this paper:

\begin{itemize}
    \item{How does the accuracy of zero-shot learning change when introducing minor variations to the original test input?}
    \item{What impact do language features, such as vocabulary overlapping, have on zero-shot learning?}
\end{itemize}

Our novel contributions are as follows:
  \begin{itemize}
    \item{We conducted four perturbations to evaluate NER models' robustness in zero-shot learning across 21 languages. The two most crucial methods included replacing named entities shared between the HRL and the LRL with entities unique to the LRL, and modifying surrounding words to assess cross-lingual adaptability.} 
    \item{We created a comprehensive section title dataset for 21 LRLs and performed two perturbations on section title prediction tasks: first, by substituting the common words between source and target languages with unique words in the target languages using the cosine similarity function, and second, by choosing substitutions randomly.}
    \item We assess the relationship between vocabulary overlap, cross-lingual transfer robustness, and adversarial perturbations.
 \end{itemize}

 Our code and novel datasets can be found at \href{https://github.com/csu-signal/xlingual-robustness}{https://github.com/csu-signal/xlingual-robustness}.


\section{Related Works}

Recent NLP models on common tasks, such as NER, have exhibited remarkable performance on well-established benchmarks with standard train/dev/test splits. Despite recent calls for alternative evaluation approaches like random splits, multiple test sets, and the introduction of a "tune-set" \cite{van2021we,gorman2019we,sogaard2021we}, these suggestions have not significantly influenced NER research practices \cite{vajjala2022we}. While some papers conduct experiments on multiple datasets \cite{bernier2020hardeval,ushio2021t}, they still primarily present results based on the standard splits of each dataset. Although a few papers acknowledge variance across multiple training runs \cite{strubell2017fast}, there is limited analysis of how model performance changes when subjected to non-standard splits or when using slightly modified test sets for NER evaluation. Even top-performing NLP systems face significant performance drops with minor input alterations across various NLP tasks \cite{gardner2020evaluating}.

Adversarial data creation for NLP mainly involves surface-level text modifications, such as inserting, deleting, or swapping words, characters, or sentences \cite{gao2018black,ribeiro2018semantically,jia2017adversarial}. Alternative strategies have been explored, such as paraphrasing \cite{iyyer2018adversarial} and generating text with semantically analogous content using neural models \cite{zhao2018generating,michel2019evaluation}. Some approaches involve human-in-the-loop interventions \cite{wallace2019trick}.

The ROCK-NER method performs adversarial attacks by replacing original entities with ones from Wikidata and then using a pre-trained masked language model like BERT \cite{devlin-etal-2019-bert} to generate context-level attacks.
Their experiments reveal a significant performance drop under perturbation, indicating that models memoriz entity patterns rather than performing true reasoning based on context.

\citet{vajjala2022we} introduce six new challenging adversarial test sets for evaluating NER, focusing on the English language. These sets are created using the Faker library\footnote{https://faker.readthedocs.io/en/master/}, and they pose distinct challenges. Models exhibit varying performance across NE categories and specific entity types, and introduce potential racial and gender biases in NER models. However, the focus on such studies on English overlooks how adversarial inputs can affect NLP for other less-studied, lower-resource languages, and so solutions to them may also privilege English and other well-studied languages.


\citet{calix2022saisiyat} performed name replacement using different languages as what has been done in \citet{vajjala2022we} for 8 foreign languages. Additionally, \citet{srinivasan2023multilingual} investigate input alterations in English, German, Hindi, emphasizing how predictions can differ with slight changes. This reveals inconsistencies in NER model robustness. For German and Hindi, combination of masking and random datasets show the most significant performance drop. 

This paper further investigates multilingual model fine-tuning and its robustness to adversarial input perturbations. We compare native LRL models to those performing cross-lingual transfer from an HRL, and examine the relationship between vocabulary overlap, cross-lingual transfer, linguistic and structural factors, and adversarial robustness.

\section{Datasets}

Our exploration focuses on 13 language pairs from a pool of 21 languages: Arabic/Farsi, Arabic/Hindi, Czech/Slovak, Dutch/Afrikaans, English/Scots, English/Welsh, French/Breton, French/Occitan, Indonesian/Malay, Italian/Sicilian, Spanish/Aragonese, Spanish/Asturian, and Spanish/Catalan.  These languages were chosen following the rationale established by \citet{nath2022generalized} for collecting loanword data: languages with a sufficiently high density for evaluation, with varying levels of expected vocabulary overlap. While \citet{nath2022generalized}'s data source is Wiktionary, we examined the WikiANN dataset \cite{pan2017cross}, a common multilingual NER dataset, and selected language pairs consisting of one language with greater resources in the data and one with fewer resources, where a substantial level of overlap in the vocabulary should be expected due to the languages having some areal (e.g., French/Breton) or close genetic (same sub-family) relationship (e.g., Czech/Slovak), or known history of borrowing at large scale (e.g., Arabic/Farsi).  See Table~\ref{titledataset}. One of these pairs---Arabic/Hindi---serves as a kind of ``control'' group; although there is a substantial amount of vocabulary shared due to borrowing, the two languages use different native scripts, and since we perform experiments over the raw text without transliteration, we expected there to be little token- or word-level overlap between these languages in terms of the raw text. In this paper, we will follow the pairwise notation L1/L2, where L1 refers to the HRL in a pair and L2, the LRL.  In the selected pairs, the HRL is often a major world or national language while the LRL is often a regional or minority language, providing an opportunity to examine where biases toward major world languages creep into NLP for minority or underrepresented languages.

WikiANN contains NER data for every language in our set, in standard BIO notation for person ({\tt PER}), location ({\tt LOC}), organization ({\tt ORG}), and miscellaneous ({\tt MISC}) categories, and so serves as the NER dataset for our experiments.

We chose NER as an experimental task as it is one of the most common NLP tasks in both research and industrial applications.\footnote{https://gradientflow.com/2021nlpsurvey/}  Another of the most common tasks is document classification, but a document classification corpus in all of the languages we evaluate is not available.\footnote{\citet{schwenk-li-2018-corpus} come closest, with 8 languages.}  We therefore approximate this task by creating a Wikipedia Section Title Prediction dataset for our languages of interest following the methodology of \citet{kakwani2020indicnlpsuite}. Section title prediction is an appropriate approximation for document classification because Wikipedia articles are usually sectioned into distinct topics regarding the subject, such as ``Early life" and ``Career'' for biographical articles, or ``Government'' and ``Economy'' for articles about states or localities, which parallel categories in document classification corpora like the Reuters Corpus dataset.

We built the Section Title corpus by crawling the Wikipedia pages corresponding to each specific language. We extracted the pages with at least 4 sections. Following this, we used the WikiExtractor tool \cite{Wikiextractor2015} to systematically extract sections along with their associated second and third-level titles from the Wikipedia pages. The dataset contains subsection text paired with four candidate titles, of which one is correct and the others are titles of other sections of the same article. We collected as many samples as possible for each language up to a limit of 100,000. Data sizes are given in Table~\ref{titledataset}.

All datasets were then divided into an 80:20 train/test split.



\begin{table}
\centering
\resizebox{.48\textwidth}{!}{
\begin{tabular}{lllll}
\hline
\textbf{HRL} & \textbf{Size} & \textbf{LRL} & \textbf{Size} & \textbf{Relationship}\\
\hline
     Arabic (ar) & 100K & Farsi (fa) & 100K & Borrowing \\
     Arabic (ar) & 100K & Hindi (hi) & 42.6K & Borrowing \\
     Czech (cs) & 100K & Slovak (sk) & 61.1K & Areal, Genetic \\
     Dutch (nl) & 100K & Afrikaans (af) & 29.7K & Genetic \\
     English (en) & 100K & Scots (sco) & 5.1K & Areal, Genetic \\
     English (en) & 100K & Welsh (cy) & 15.2K & Areal, Borrowing \\
     French (fr) & 100K & Breton (br) & 8.1K & Areal, Borrowing \\
     French (fr) & 100K & Occitan (oc) & 13.7K & Areal, Genetic \\
     Indonesian (id) & 100K & Malay (ms) & 60.3K & Areal, Genetic \\
     Italian (it) & 100K & Sicilian (scn) & 1.4K & Areal, Genetic \\
     Spanish (es) & 100K & Aragonese (an) & 5.1K & Areal, Genetic \\
     Spanish (es) & 100K & Asturian (ast) & 85.5K & Areal, Genetic \\
     Spanish (es) & 100K & Catalan (ca) & 100K & Areal, Genetic \\
\hline
\end{tabular}}
\caption{\label{titledataset}
Size of languages for section title prediction dataset, and relationship between languages in studied pair.
}
\end{table}

\section{Methodology}
\label{sec:method}

To assess the effect of zero-shot transfer between languages with overlapping vocabulary, we compare the performance of the MBERT and XLM-R models. These are two of the most well-known multilingual, publicly-available encoder-style models in use, notable for their abilities to align semantically similar representations across languages and for their multilingual task performance \cite{pires2019multilingual,conneau2020unsupervised}. We evaluate both models in a native setting when they are fully fine-tuned on the two tasks in an LRL; in a transfer setting, where they are trained on an HRL and evaluated on the paired LRL; and under different perturbations of the data. Details are given in Sec.~\ref{sec:exp}.

The two tasks chosen represent two fundamental NLP inference challenges: information extraction (IE) from unstructured texts, encompassing the identification of individuals' names, organizations, geographical locations, etc.; and the selection of the appropriate classification of a text from multiple options, requiring the selection of the most appropriate title for a section text among the four presented choices. NER is a valuable IE task to assess the effects of vocabulary overlap on cross-lingual transfer, because named entities in LRLs are often borrowed from HRLs with minimal modifications.  Title selection is a valuable classification task for similar reasons: section texts represent ``documents'' where key words evidencing the section title options may be shared or similar between languages.  For instance in the Welsh example ``{\it mae logo'r ddarpar fanc},'' {\it logo} overlaps with English and predicts the section title ``logo.''

\subsection{Perturbation methods}
\label{ssec:perturb}
Two additional small datasets were gathered for the perturbation process:
\begin{enumerate*}[label=\arabic*)]
    \item A dataset of given names for each target language scraped from the {\tt [Language]\_given\_names} category of Wiktionary.
    \item A dataset of places for each target language scraped from its {\tt Places} category in Wiktionary.
\end{enumerate*}

We implemented four methods to generate adversarial sets:

\begin{enumerate}
    \item Change given names (first element) of all PER entities to randomly-chosen elements of the given names dataset in the same language.  

    \item Change names of all LOC entities to randomly-chosen elements of the placenames dataset in the same language.
    
    \item Replace named entities in the L2 test file that also occur in the L1 training file with a named entity with the same tag that occurs in L2 test but not in L1.
    
    For example, {\it Tour Eiffel} (Eiffel Tower) is the same in Breton and French, so it may be replaced with {\it Bolz-enor Pariz} (Arc de Triomphe), which is the same NER type, but non-overlapping.
    
    \item Leave the the named entity unchanged, but instead take {\it surrounding} words in the L2 test file that occur in the L1 training file and replace them with words unique to L2 test that are not punctuation or stop words.\footnote{\href{https://github.com/stopwords-iso/stopwords-iso/}{https://github.com/stopwords-iso/stopwords-iso/} provided the stop word lists.} The substitute word is the word with the highest cosine similarity with the original word.\footnote{Computing the most similar word follows \citet{nath2022generalized} and constructs a dummy ``sentence'' consisting of {\tt [CLS]<word>[SEP]} (MBERT) or {\tt <bos><word><eos>} (XLM-R) for each word, and computing the cosine distance between the contextualized first token representation.}
    
    For example, in {\it An Tour Eiffel zo un tour metalek savet e Pariz gant Gustave Eiffel} (``The Eiffel Tower is a metal tower built in Paris by Gustave Eiffel''), the word {\it tour} is the same in Breton and French, so it is replaced with a semantically-similar Breton word that doesn't also occur in the French wordlist. Table \ref{wordexample} shows a sample of original words and their substitutes determined by cosine similarity according to MBERT and XLM-R. This sample shows how bias can creep into the substitutions, {\it viz.} ``males'' for ``hijackers''.
    
    \item Combine the perturbations 3 and 4 to change the entities and surrounding words at the same time.
\end{enumerate}

For section title prediction, we use only perturbation 4, which does not require the training data to be tagged with NE labels.

Adversarial set generation was conducted automatically. For semantic-level perturbations like perturbation 4, a manually-created semantic resource like WordNet is appealing, but infeasible due to some of the languages we examine, for which there do not exist sufficient WordNet or WordNet-like resources.  For instance, we considered Global WordNet\footnote{\href{https://omwn.org/}{https://omwn.org/}} but even this resource does not exist for all of our languages of interest, for instance Aragonese.  This is to be expected given that the domain in question is low-resource languages. Secondly, WordNets that do exist for the languages of interest are incomplete or very small, lacking a substantial number of words. Thirdly, WordNet synsets do not necessarily cover alternative potential substitutions that would be sensical, without traversing the synset tree quite far from the word of origin.  For instance, instead of saying ``I see my sister that day," a reasonable perturbation would be ``I see my mother that day," but despite being semantically similary, ``sister'' and ``mother'' are not in a synset. Similarly, a WordNet-based approach for synonym replacement becomes challenging when dealing with homographs, such as (in English) {\it might}/{\it could} vs. {\it might}/{\it strength}. Therefore, for consistency across all language, we used the automated method as described.

\begin{table}
\centering
\resizebox{.48\textwidth}{!}{
\begin{tabular}{lll}
\hline
\textbf{Content word} & \textbf{MBERT option} & \textbf{XLM-R option}\\
     \hline
     channels & shots & broadcasts \\
     bred & lived & assistant \\
     population & parted & people \\
     serve & carried & arrangement \\
     place & event & there \\
     journalist & lawyer & activist \\
     female & woman & woman \\
     hijackers & triumphs & males \\
     defeated & won & defeating \\
     \hline
\end{tabular}}
\caption{\label{wordexample}
Sample of highest cosine-similarity alternatives existing in the test split of the English dataset. Since we focus on non-English languages, these are note actual examples from the data but rather illustrative of the phenomenon and extracted using the same methdology in use.}
\end{table}

\subsection{Computing vocabulary overlap}


To investigate the correlation between vocabulary overlap and zero-shot knowledge transfer across languages, we started by extracting all labeled NER chunks within the datasets of the paired languages, and computing the percentage of identical words with identical labels---excluding those tagged {\tt O}.  For a pair, the percentage of overlap between L1 and L2 is considered to be number of words shared between the L1 training set and L2 test set divided by the total number of words in the L2 test set. See Table~\ref{ner-overlap}.

For section title prediction, we identified common and unique words for perturbation, and performed overlap computation, using the first 128 tokens from each section. Due to variances in tokenization between MBERT and XLM-R, there may be different values for overlap between the two models (see Table~\ref{title-overlap}).

\begin{table}
\centering
\begin{tabular}{lll}
\hline
\textbf{L1} & \textbf{L2} & \textbf{\% overlap}\\
\hline
        ar & hi & 4.88 \\ 
        ar & fa & 19.94 \\ 
        cs & sk & 39.55\\ 
        nl & af & 31.57 \\ 
        en & sco & 25.19 \\ 
        en & cy & 22.07 \\ 
        fr & br & 23.33 \\ 
        fr & oc & 23.61 \\ 
        it & scn & 43.17 \\ 
        id & ms & 41.87 \\ 
        es & an & 46.26 \\ 
        es & ast & 47.66 \\ 
        es & ca & 36.77 \\ 
\hline
\end{tabular}
\caption{\label{ner-overlap}
Named entity overlap in L1-train/L2-test for NER.
}
\end{table}

\begin{table}
\centering
\begin{tabular}{llll}
\hline
\textbf{L1} & \textbf{L2} & \textbf{Model} & \textbf{\% overlap}\\
\hline
     ar & hi & MBERT & 2.12 \\ 
     ar & hi & XLM-R & 1.98 \\
     ar & fa & MBERT & 14.65 \\ 
     ar & fa & XLM-R & 15.01 \\
     cs & sk & MBERT & 24.26 \\ 
     cs & sk & XLM-R & 24.18 \\
     nl & af & MBERT & 22.63 \\ 
     nl & af & XLM-R & 22.57 \\
     en & sco & MBERT & 29.22 \\ 
     en & sco & XLM-R & 29.19 \\
     en & cy & MBERT & 17.31 \\ 
     en & cy & XLM-R & 17.08 \\
     fr & br & MBERT & 9.50 \\ 
     fr & br & XLM-R & 9.44 \\ 
     fr & oc & MBERT & 23.09 \\ 
     fr & oc & XLM-R & 23.04 \\
     id & ms & MBERT & 36.34 \\ 
     id & ms & XLM-R & 36.34 \\
     it & scn & MBERT & 25.99 \\ 
     it & scn & XLM-R & 25.86 \\
     es & an & MBERT & 24.80 \\ 
     es & an & XLM-R & 24.77 \\ 
     es & ast & MBERT & 29.59 \\ 
     es & ast & XLM-R & 29.65 \\
     es & ca & MBERT & 17.12 \\ 
     es & ca & XLM-R & 17.20 \\
\hline
\end{tabular}
\caption{\label{title-overlap}
Word overlap in L1-train/L2-test for title prediction task.
}
\end{table}



\section{Evaluation}
\label{sec:exp}

We used the {\tt bert-base-multilingual-cased} \cite{devlin-etal-2019-bert} (MBERT) and {\tt xlm-roberta-base} \cite{conneau2020unsupervised} (XLM-R) variants.  We fine-tuned each of these models in the two tasks in multiple conditions:

\begin{enumerate*}[label=\arabic*)]
    \item {\it Native L2}: A standard fine-tuning on the L2 training data and testing on the L2 test data;
    \item {\it Cross-lingual transfer}: Fine-tuning on the L1 training data and testing on the L2 test data. Since the language pairs were selected for vocabulary overlap, this condition allowed us to assess the level to which performance on a LRL can be achieved by exposure to data only from an HRL that may contain similar task-relevant vocabulary.
    \item {\it Perturbation}: In each of the two above conditions, the task-relevant perturbations are applied, to further assess the extent to which cross-lingual transfer or native performance is robust to adversarial changes to the input.
\end{enumerate*}

To account for randomness in training and testing sample selection, which could lead to disparate values, we averaged the results across three runs.  Our primary evaluation metric is F$_1$ score relative to token overlap in the chosen sentences rather than the entire token pool. This methodology aligns with our earlier strategy, where we concentrated on examining the overlap between the L1 training set and the L2 testing set, rather than the entire datasets of L1 and L2, offering a more precise insight into the multilingual capacities of the models relative to specific training and testing data.

\section{Results}

\begin{table*}
\centering
\resizebox{\textwidth}{!}{
\begin{tabular}{lllllllllllllllllllll}
\hline
&& \multicolumn{9}{c}{\textbf{MBERT}} && \multicolumn{9}{c}{\textbf{XLM-R}} \\
            \cline{3-11} \cline{13-21}
&& \multicolumn{7}{c}{\textbf{NER}} & \multicolumn{2}{c}{\textbf{WikiTitle}} && \multicolumn{7}{c}{\textbf{NER}} & \multicolumn{2}{c}{\textbf{WikiTitle}} \\
            \cline{3-8} \cline{10-11} \cline{13-18} \cline{20-21}
\textbf{Train} & \textbf{Test} & \textbf{Base} & \textbf{P1} & \textbf{P2} & \textbf{P3} & \textbf{P4} & \textbf{P5} && \textbf{Base} & \textbf{P4} && \textbf{Base} & \textbf{P1} & \textbf{P2} & \textbf{P3} & \textbf{P4} & \textbf{P5} && \textbf{Base} & \textbf{P4} \\
\hline
ar & hi & 67.2 & 64.2 & 68.9 & 67.2 & 67.2 & 67.2 && 63.6 & 63.0 && 67.3 & 67.4 & 70.7 & 67.3 & 67.3 & 67.3 && 75.8 & 75.0 \\
hi & hi & 86.7 & 86.5 & 87.2 & 71.3 & 79.0 & 66.7 && 73.8 & 72.5 && 87.5 & 87.2 & 88.1 & 76.6 & 80.7 & 68.3 && 77.8 & 77.1 \\
\hline
ar & fa & 45.0 & 43.0 & 44.7 & 45.0 & 45.0 & 44.9 && 79.3 & 77.1 && 43.6 & 42.8 & 40.1 & 43.6 & 43.5 & 43.4 && 78.0 & 73.9 \\
fa & fa & 90.3 & 88.0 & 89.1 & 86.5 & 60.8 & 56.7 && 81.6 & 79.1 && 89.4 & 88.2 & 87.4 & 85.5 & 78.2 & 74.1 && 81.0 & 76.5 \\
\hline
cs & sk & 82.9 & 82.4 & 87.0 & 78.4 & 82.5 & 77.9 && 80.3 & 75.6 && 78.0 & 77.2 & 86.1 & 73.4 & 78.1 & 73.5 && 80.3 & 73.3 \\
sk & sk & 92.6 & 91.7 & 91.0 & 86.4 & 92.1 & 85.0 && 83.5 & 78.5 && 91.5 & 91.1 & 89.8 & 81.5 & 88.6 & 77.5 && 82.3 & 75.1 \\
\hline
nl & af & 81.2 & 81.0 & 83.8 & 78.4 & 81.2 & 78.6 && 78.5 & 71.6 && 79.9 & 80.0 & 81.5 & 77.8 & 79.3 & 76.9 && 75.4 & 71.6 \\
af & af & 92.2 & 91.6 & 92.1 & 81.1 & 89.5 & 78.5 && 81.3 & 74.3 && 89.8 & 90.0 & 90.8 & 77.9 & 86.2 & 76.0 && 76.8 & 66.9 \\
\hline
en & sco & 78.3 & 77.9 & 72.0 & 71.0 & 78.2 & 71.7 && 85.7 & 76.2 && 62.4 & 62.0 & 60.6 & 60.6 & 63.2 & 61.3 && 75.5 & 62.5 \\
sco & sco & 93.4 & 93.0 & 83.2 & 81.0 & 91.4 & 79.2 && 88.6 & 80.8 && 90.2 & 89.6 & 82.5 & 79.6 & 87.5 & 75.0 && 71.5 & 60.2 \\
\hline
en & cy & 62.5 & 61.8 & 65.3 & 61.3 & 62.4 & 61.6 && 67.5 & 63.6 && 61.5 & 61.2 & 64.9 & 60.4 & 61.4 & 60.4 && 61.7 & 58.8 \\
cy & cy & 92.6 & 91.9 & 87.1 & 77.0 & 89.5 & 75.0 && 76.6 & 73.5 && 90.9 & 90.4 & 85.1 & 76.1 & 83.1 & 67.8 && 72.1 & 67.3 \\
\hline
fr & br & 74.3 & 71.8 & 73.5 & 73.3 & 74.2 & 72.8 && 66.6 & 63.1 && 66.3 & 64.2 & 66.6 & 64.7 & 66.3 & 64.5 && 59.3 & 54.0 \\
br & br & 92.8 & 88.4 & 88.2 & 84.5 & 88.8 & 79.9 && 71.1 & 66.1 && 89.1 & 85.8 & 87.1 & 81.3 & 82.8 & 74.1 && 59.3 & 55.2 \\
\hline
fr & oc & 83.9 & 83.7 & 89.1 & 83.5 & 83.7 & 83.4 && 76.6 & 71.9 && 72.5 & 72.3 & 78.8 & 71.8 & 72.3 & 71.9 && 66.5 & 59.1 \\
oc & oc & 95.3 & 94.9 & 95.8 & 92.3 & 87.8 & 83.9 && 79.1 & 75.2 && 93.8 & 93.0 & 94.6 & 91.5 & 92.6 & 89.8 && 67.0 & 61.3 \\
\hline
id & ms & 68.7 & 67.7 & 76.7 & 64.8 & 68.5 & 64.8 && 79.9 & 68.4 && 69.7 & 69.5 & 79.9 & 66.2 & 69.5 & 65.8 && 78.3 & 58.4 \\
ms & ms & 92.4 & 92.6 & 83.5 & 81.7 & 81.8 & 70.5 && 82.7 & 71.8 && 92.4 & 91.9 & 89.1 & 71.7 & 79.7 & 59.5 && 80.3 & 62.4 \\
\hline
it & scn & 63.7 & 63.3 & 80.2 & 58.4 & 49.5 & 45.4 && 71.0 & 66.2 && 60.8 & 60.7 & 74.0 & 55.3 & 50.4 & 45.5 && 60.7 & 46.8 \\
scn & scn & 92.9 & 91.1 & 88.1 & 79.8 & 74.4 & 64.9 && 64.3 & 57.1 && 90.5 & 88.2 & 82.8 & 79.7 & 72.4 & 62.5 && 40.0 & 39.0 \\
\hline
es & an & \textbf{88.0} & 87.9 & 84.8 & 85.4 & 80.7 & 77.5 && 86.1 & 76.3 && \textbf{86.1} & 86.2 & 86.4 & 83.3 & 75.3 & 72.9 && 77.0 & 55.0 \\
an & an & 95.8 & 95.8 & 88.4 & 85.6 & 90.9 & \textbf{79.1} && 83.4 & 76.8 && 94.2 & 93.6 & 92.5 & 79.8 & 80.4 & \textbf{66.1} && 72.6 & 59.4 \\
\hline
es & ast & \textbf{90.4} & 90.2 & 86.0 & 85.1 & 89.6 & 84.6 && 84.1 & 77.5 && \textbf{84.3} & 84.2 & 86.0 & 77.0 & 84.1 & 76.3 && 76.7 & 59.6 \\
ast & ast & 93.6 & 92.8 & 90.1 & 82.7 & 93.3 & \textbf{79.7} && 85.2 & 78.4 && 89.6 & 89.2 & 90.1 & 77.7 & 90.0 & \textbf{76.4} && 80.3 & 68.0 \\
\hline
es & ca & \textbf{85.1} & 84.3 & 87.2 & 84.0 & 85.1 & 84.0 && 79.3 & 75.9 && \textbf{82.6} & 82.8 & 83.9 & 80.8 & 82.3 & 79.8 && 72.8 & 66.2 \\
ca & ca & 92.3 & 91.5 & 91.6 & 87.3 & 91.6 & \textbf{86.5} && 85.9 & 83.0 && 89.4 & 89.6 & 88.0 & 83.3 & 88.6 & \textbf{82.1} && 83.9 & 78.0 \\
\hline
\end{tabular}}
\caption{\label{big-results}
F$_1\times$100 (NER) and accuracy (title prediction) scores for MBERT and XLM-R without perturbation ({\bf Base}) and with all applicable perturbations on all evaluated language pairs.  P1-5 references the different perturbations described in the list in Sec.~\ref{ssec:perturb}. Bold numbers refer to native and cross-lingual NER accuracy values when the source language is Spanish, which are discussed further as noteworthy cases in Sec.~\ref{sec:disc}.
}
\end{table*}

Table~\ref{big-results} shows the performance on NER and title selection of MBERT and XLM-R, for all evaluated language pairs, with baseline (unperturbed) scores, and scores after all applicable perturbations.  Table~\ref{p-values} shows the statistical significance of the performance changes associated with each perturbation, given the model and cross-lingual transfer setting.

We can see that using an HRL\textrightarrow LRL transfer setting never reaches the performance of the native LRL fine-tuning, falling below by $\sim$1-30\% F$_1$/accuracy. Where cross-lingual transfer comes closest is in language pairs that are geographically close and genetically close (e.g., Spanish/Asturian), because core vocabulary is likely to be similar already, and the document sets in the training data likely share named entities like names of people and locations that are commonly discussed in the two languages.  Interestingly, though, we see that the cross-lingual transfer models appear to be more robust to certain perturbations, such as P4 (perturbing context words), which by itself did not significantly change the NER results for MBERT or XLM-R using cross-lingual transfer.  MBERT cross-lingual models are also more robust to P2 than XLM-R cross-lingual models, as the perturbation of LOC tags also did not affect results to a statistically significant extent.  XLM-R was more robust to random replacement of B-PER tags.  On average, MBERT appears more robust to the perturbations we applied, where even the performance changes that were statistically significant were less so than those of XLM-R.  However, we should note that even the simple perturbation of changing context words in the title selection task degraded performance to a very significant level nearly across the board.

\begin{table*}
\centering
\resizebox{\textwidth}{!}{
\begin{tabular}{llllllllll}
& \multicolumn{3}{c}{\textbf{MBERT}} && \multicolumn{3}{c}{\textbf{XLM-R}}  \\
\hline
& NER: L2 & avg. $\Delta$ F$_1$ & NER: L1\textrightarrow L2 & avg. $\Delta$ F$_1$ && NER: L2 & avg. $\Delta$ F$_1$ & NER: L1\textrightarrow L2 & avg. $\Delta$ F$_1$ \\
\hline
P1 & $p = 0.0118$ & -1.00 & $p = 0.0046$ & -0.92 && $p = 0.0116$ & -0.80 & $p = 0.0655$ & -0.34 \\
P2 & $p = 0.0033$ & -3.65 & $p = 0.2096$ & 2.15 && $p = 0.0165$ & -2.33 & $p = 0.0246$ & 3.42 \\
P3 & $p < 0.0001$ & -9.66 & $p = 0.0013$ & -2.72 && $p < 0.0001$ & -10.46 & $p = 0.0013$ & -2.52 \\
P4 & $p = 0.0105$ & -7.07 & $p = 0.1500$ & -1.80 && $p = 0.0004$ & -6.73 & $p = 0.1499$ & -1.69 \\
P5 & $p < 0.0001$ & -16.71 & $p = 0.0106$ & -4.36 && $p < 0.0001$ & -17.62 & $p = 0.0090$ & -4.26 \\
\hline
& Titles: L2 & avg. $\Delta$ acc. & Titles: L1\textrightarrow L2 & avg. $\Delta$ acc. && Titles: L2 & avg. $\Delta$ acc. & Titles: L1\textrightarrow L2 & avg. $\Delta$ acc. \\
\hline
P4 & $p < 0.0001$ & -5.38 & $p < 0.0001$ & -5.54 && $p = 0.0002$ & -7.57 & $p = 0.0003$ & -9.52 \\
\hline
\end{tabular}}
\caption{\label{p-values}
Effects of different perturbations, per model type by a paired, two-tailed $t$-test, and average change in F1/accuracy. The average change in F1/accuracy metrics after perturbation appears significantly less during cross-lingual transfer than in the native setting. While XLM-R demonstrates nearly equivalent robustness to perturbation in both settings in NER when compared to MBERT, its robustness diminishes in the sentence-level task---section title prediction---where word memorization might be more applicable.}

\end{table*}

\section{Discussion}
\label{sec:disc}

\subsection{NER}

\begin{figure}[!ht]
\begin{center}
\includegraphics[width=.5\textwidth]{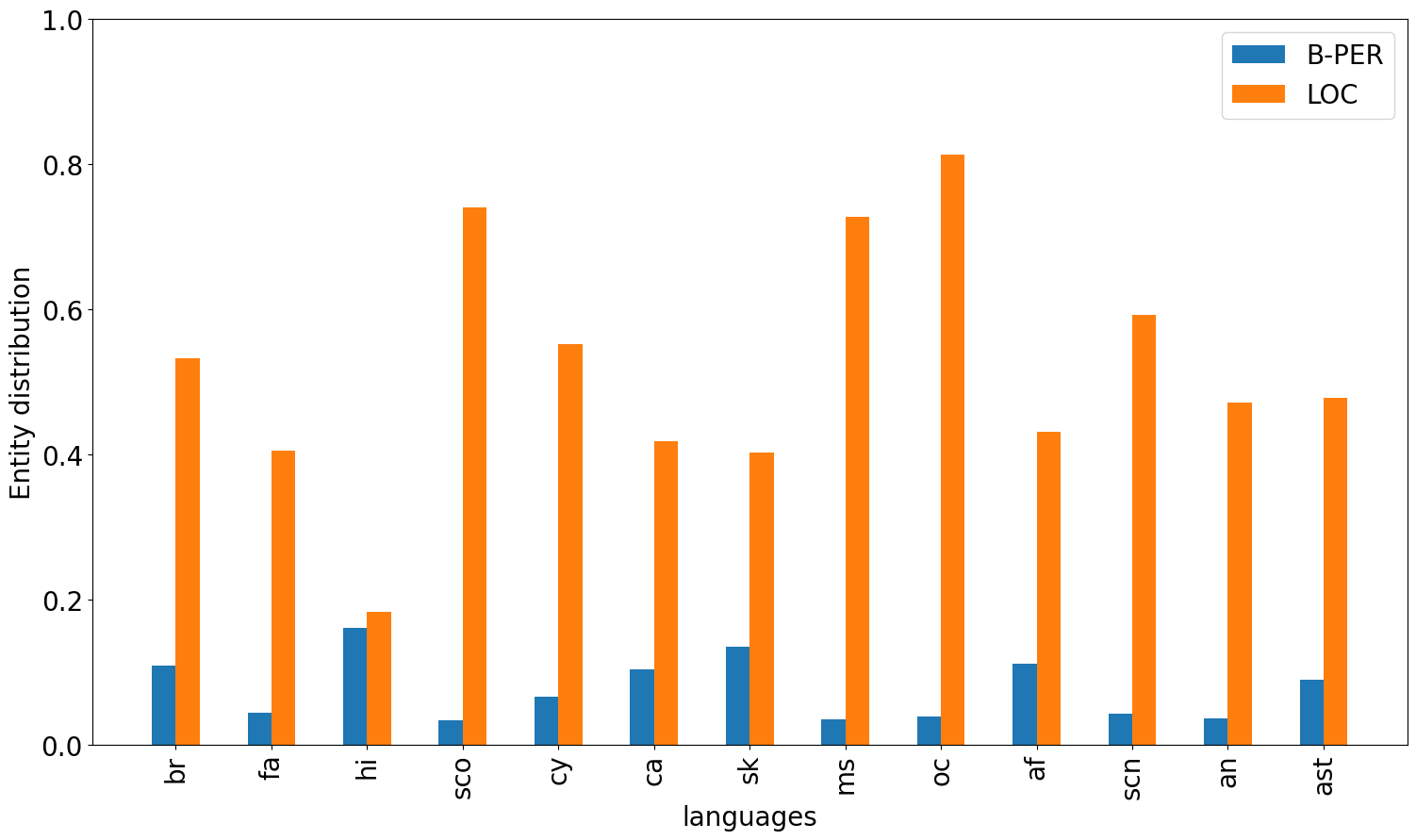} 

\caption{WikiANN distribution of B-PER and LOC for different LRLs.}
\label{per-loc-dist}
\end{center}
\end{figure}

Under perturbation of {\tt B-PER} tokens (P1), macro F$_1$ score changes between 0--4\% and F$1$ of {\tt PER} classes changes from 1--13\%, depending on the {\tt B-PER} distribution and the number of available alternatives.  Under perturbation of {\tt LOC} tokens (P2), macro F$_1$ score changes from 1--13\% and {\tt LOC} F$_1$ (averaged across {\tt B-LOC} and {\tt I-LOC}) changes from 1--27\%.  Although {\tt LOC} tokens form a greater proportion of the test data than {\tt B-PER} tokens alone, perturbing the {\tt LOC} tokens causes far less drop in performance.  One reason may be that many {\it LOC} entities in the test sets include 3--6 individual tokens, while the alternate candidates scraped from Wiktionary mostly include 2 tokens, making it easier to segment shorter NE chunks.  Notably, {\tt LOC} perturbation frequently causes performance on a language pair to {\it rise}, perhaps significantly (see Italian/Sicilian), signaling cases where cross-lingual transfer provides increased robustness to adversarial data, relative to baseline performance. Distributions of {\tt B-PER} and {\tt LOC} tokens in the LRL test sets are shown in Fig.~\ref{per-loc-dist}.

\begin{figure*}[!ht]
\begin{center}
\includegraphics[width=\textwidth]{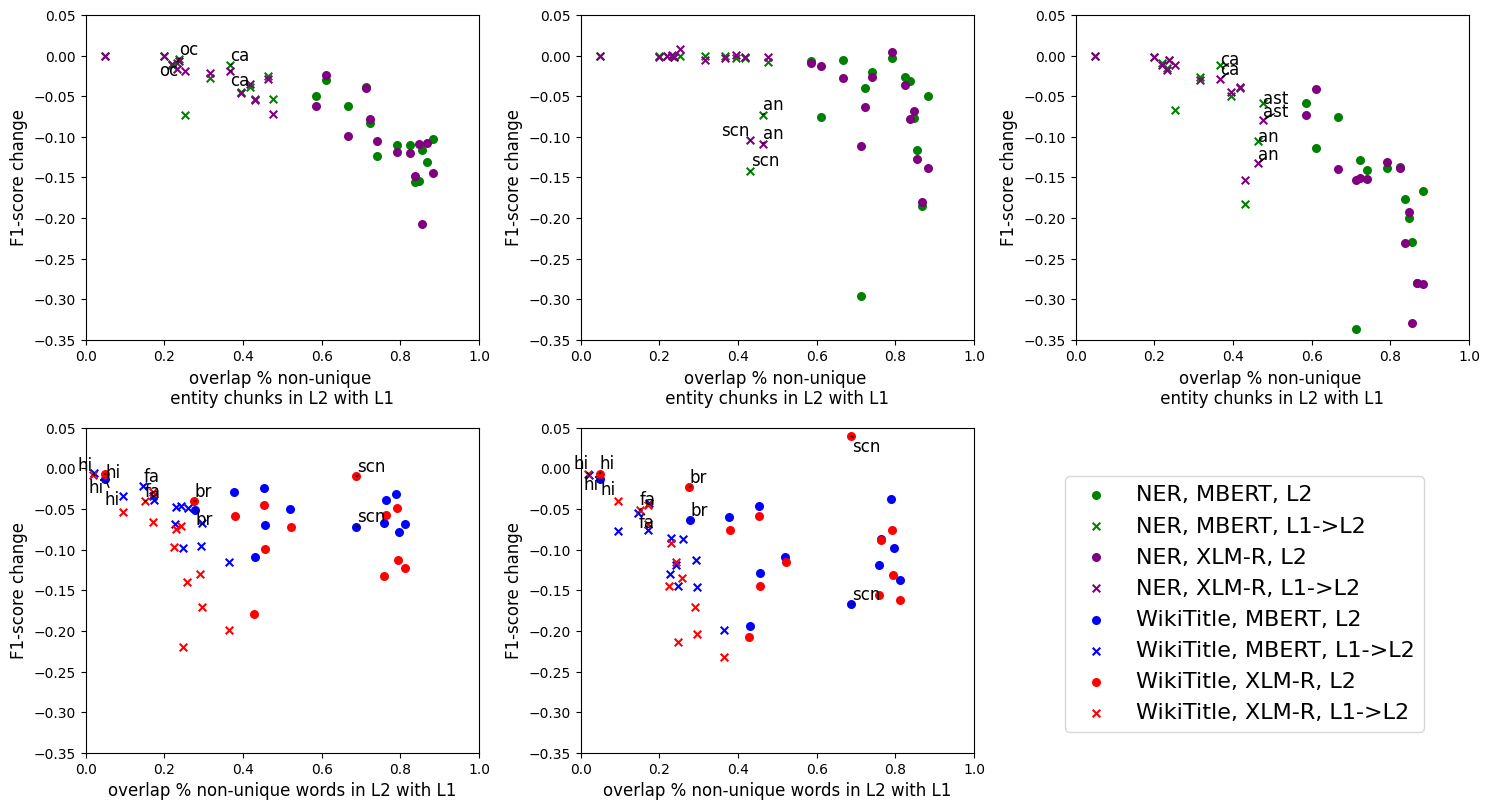} 


\caption{Change in F$_1$ score under perturbation as a function of degree of vocabulary overlap. Left to right, top to bottom: P3 for NER, P4 for NER, P5 for NER, P4 for title selection, P4 for title selection using random substitutions instead of the most cosine-similar words.}
\label{points-all-plots}
\end{center}
\end{figure*}

Fig.~\ref{points-all-plots} (first row) shows macro F$_1$ change under perturbations 3, 4, and 5 as a function of the degree of vocabulary overlap between L1 and L2 for all pairs. We observe a clear correlation between the proportion of shared vocabulary items between the train and test sets and the performance degradation when test entities are perturbed to substitute those that are shared between the sets with unique entities.  In the cross-lingual transfer models, this removes words that are common between L1 and L2 and replaces them with words unique to L2.  This suggests that multilingual models' NER performance for LRLs depends to some extent on word memorization, and the extent to which this is true is a function of vocabulary overlap with other, more well-resourced languages; the model may not be recognizing that a term is a named entity in Occitan or Catalan, but rather one from a French or Spanish corpus and is ``riding'' its ability to perform in those languages.

Degradation of performance under P4 (changing surrounding context words---see Fig.~\ref{points-all-plots}, top center) is pronounced in the native L2 models, but for most cross-lingual transfer models, degradation is small.  Two notable exceptions are Italian/Sicilian and Spanish/Aragonese, where perturbing context words causes a drop of $\sim$8--20 F$_1$ points.

Under perturbation 5 (the combination of perturbations 3 and 4), NER performance suffers a fairly precipitous drop. The three pairs involving Spanish (Spanish/Aragonese, Spanish/Asturian, and Spanish/Catalan; bolded in Table~\ref{big-results}) are notable here, in that P5 brings the native model down to the performance level of the unperturbed cross-lingual transfer model (in the case of Aragonese and Asturian, perturbed native model performance drops {\it below} the performance of the model trained for cross-lingual transfer from Spanish). This also suggests that on these LRLs, MLLMs may be leveraging their capabilities in Spanish to achieve their initial performances.



\subsection{Section Title Prediction}
\label{ssec:stp}

In this task as with NER, in the case of cross-lingual transfer, the degree of overlap has a significant impact on the F$_1$ score, with a noticeable drop in F$_1$  score when key words in the target language are removed.

In the lower center of Fig.~\ref{points-all-plots}, the plot shows the effect of perturbation 4 (perturbing context words) on the title selection, when the substitutes are chosen randomly instead by choosing the most similar candidate according to the cosine function. We can see that MBERT suffers more from the random perturbation than from the cosine perturbation in the cross-lingual transfer condition, but both models suffer more from this perturbation when compared to the other one in the native condition.

One point of note is that in the case of Arabic/Hindi, the one pair where the two languages use different native scripts, none of the perturbations appeared to have much effect in the cross-lingual setting.  This is expected, due to the low default token overlap: a model fine-tuned on Arabic will have difficulty handling Hindi words written in Devanagari, regardless of what they are.  What is interesting is that in the case of Arabic/{\it Persian}, which do share the same script, the same is true, and Arabic/Persian cross-lingual transfer performance on NER is substantially lower than on Arabic/Hindi, despite the differences in script and therefore tokenization.

A native Persian speaker conducted a qualitative analysis of words judged similar for use in P4 and P5, according to their cosine similarity. In both MBERT and XLM, similar words were often found to rhyme or share subwords: e.g., {\it m\^{a}rk} (``brand'') vs. {\it m\^{a}rd} (``evil''), or {\it sard} (``story'') vs. {\it sardard} (``headache''). This implies that subword tokens are being overvalued when computing vectors from {\tt [CLS]/<bos>} tokens in Persian, and perhaps other LRLs.

In the native condition, overlap is computed using non-stop words found in both the training and test files.  Consequently, when the value is low, as is the case with Breton, we would expect performance under perturbation to remain relatively unchanged (compare Hindi), but Breton still suffers a performance loss of $\sim$4--5 F$_1$ points. This suggests that this task relies heavily on word memorization of the training data, as a similar drop in performance is observed when words are substituted randomly. The semantic similarities of the substitute words under P4 seem to not matter. Sicilian performance in MBERT substantially exceeds that of XLM-R, but also suffers more under perturbation. Sicilian training data is included in the pretraining data for MBERT but not for XLM-R, which partially explains this trend, but the much lower performance of the native Sicilian XLM-R model on title selection compared to NER suggests that NER fine-tuning can leverage other representations (e.g., common named entities between Italian and Sicilian) in a way that a task that requires inference over more common words, like title selection, cannot.



\section{Conclusion and Future Work}

In this paper, we have presented a set of adversarial perturbations to test the ability of language models to generalize from higher-resourced languages to lower-resourced languages in a cross-lingual transfer zero-shot setting.  Our experiments are performed in a language-agnostic manner for both NER and title selection tasks.  To our knowledge, this is the first time such an experimental set has been performed with an explicit focus on LRLs and cross-lingual transfer from HRLs.  We conducted evaluations on 21 languages, encompassing both high and low-resource languages, employing two widely recognized multilingual models, MBERT and XLM-R. Results exhibit variations across different languages, influenced by their linguistic structures and similarities. Our core findings can be summarized as follows:

\begin{itemize}
    \item There is a pronounced effect of vocabulary overlap on NER performance. Perturbing named entities so that the test data contains only non-overlapping words has a statistically very significant impact on model performance.
    \item Although models utilizing cross-lingual transfer typically exhibit lower numerical performance than models trained in a native LRL setting, they are often somewhat more robust to certain types of perturbations of the input.
    \item Title selection, as a proxy for document classification, in LRLs appears to heavily rely on word memorization.
\end{itemize}

These proposed test sets have the potential for further exploration, particularly in challenging tokenizers directly. For example, the Persian examples discussed in Sec.~\ref{ssec:stp} suggest that, although BPE tokenization methods should help LRL performance by not biasing token vocabulary toward frequent tokens in a specific language \cite{sennrich2016neural}, similarity between subword tokens may be overvalued when optimizing the embedding space.  This, among other factors, motivates the need for an equitable consideration of lower-resource languages in building NLP models \cite{joshi2020state}.



Our results show that MLLMs exhibit some level of ``mutual intelligibility'' between individual languages; a model can ``get by" in a language like Asturian if it knows a similar language like Spanish.  However, a risk of MLLMs is a systematic encoding of biases toward HRLs, which also has implications for representations of minority and regional languages. Our own datasets, due to availability of online resources, are still biased toward Indo-European languages.  A multilingual capacity is not necessarily enough to handle an arbitrary input language \cite{virtanen2019multilingual,scheible2020gottbert,tanvir2021estbert,nath-etal-2023-axomiyaberta}, and performance is sensitive to minor changes to the input, even without perturbations in the latent space \cite{narasimhan-etal-2022-towards}.

Finally, this research has been conducted on encoder models. The reasons for this are manifold but center around the fact that encoder models, being older and smaller, typically demand fewer computational resources, allowing us to perform more experiments. Additionally, unlike currently-touted SOTA decoder models like GPT-4, most encoder model weights and processing pipelines are freely available on platforms like HuggingFace \cite{wolf2019huggingface}, meaning that we can directly access the embedding spaces to inform our perturbation techniques. Nonetheless, our findings could influence directions for probing and prompt engineering for generative models that exhibit multilingual capability. Most open-weight generative models (e.g., LLaMA 2 \cite{touvron2023llama}) are not multilingual; those that are, such as ChatGPT/GPT-4 \cite{qin2023chatgpt}, remain closed.  However, since our techniques are general, they could be applied to open-source multilingual generative models like XGLM \cite{lin2021few}. We do note that multilingual generative models still do not necessarily all the languages we study here, which further indicates a resource deficiency for many languages when it comes to SOTA generative NLP.

%
\section*{Bibliographical References}\label{reference}

\bibliographystyle{lrec-coling2024-natbib}
\bibliography{lrec-coling2024-example}

\begin{thebibliography}{40}
\expandafter\ifx\csname natexlab\endcsname\relax\def\natexlab#1{#1}\fi

\bibitem[{Attardi(2015)}]{Wikiextractor2015}
Giuseppe Attardi. 2015.
\newblock Wikiextractor.
\newblock \url{https://github.com/attardi/wikiextractor}.

\bibitem[{Bernier-Colborne and Langlais(2020)}]{bernier2020hardeval}
Gabriel Bernier-Colborne and Philippe Langlais. 2020.
\newblock Hardeval: Focusing on challenging tokens to assess robustness of ner.
\newblock In \emph{Proceedings of the Twelfth Language Resources and Evaluation
  Conference}, pages 1704--1711.

\bibitem[{Calix et~al.(2022)Calix, Ben-Joseph, Lopatina, Ashley, Gogia,
  Sieniawski, and Brennen}]{calix2022saisiyat}
Ricardo~A Calix, Jj~Ben-Joseph, Nina Lopatina, Ryan Ashley, Mona Gogia, George
  Sieniawski, and Andrea Brennen. 2022.
\newblock Saisiyat is where it is at! insights into backdoors and debiasing of
  cross lingual transformers for named entity recognition.
\newblock In \emph{2022 IEEE International Conference on Big Data (Big Data)},
  pages 2940--2949. IEEE.

\bibitem[{Conneau et~al.(2020)Conneau, Khandelwal, Goyal, Chaudhary, Wenzek,
  Guzm{\'a}n, Grave, Ott, Zettlemoyer, and Stoyanov}]{conneau2020unsupervised}
Alexis Conneau, Kartikay Khandelwal, Naman Goyal, Vishrav Chaudhary, Guillaume
  Wenzek, Francisco Guzm{\'a}n, {\'E}douard Grave, Myle Ott, Luke Zettlemoyer,
  and Veselin Stoyanov. 2020.
\newblock Unsupervised cross-lingual representation learning at scale.
\newblock In \emph{Proceedings of the 58th Annual Meeting of the Association
  for Computational Linguistics}, pages 8440--8451.

\bibitem[{Devlin et~al.(2019)Devlin, Chang, Lee, and
  Toutanova}]{devlin-etal-2019-bert}
Jacob Devlin, Ming-Wei Chang, Kenton Lee, and Kristina Toutanova. 2019.
\newblock \href {https://doi.org/10.18653/v1/N19-1423} {{BERT}: Pre-training of
  deep bidirectional transformers for language understanding}.
\newblock In \emph{Proceedings of the 2019 Conference of the North {A}merican
  Chapter of the Association for Computational Linguistics: Human Language
  Technologies, Volume 1 (Long and Short Papers)}, pages 4171--4186,
  Minneapolis, Minnesota. Association for Computational Linguistics.

\bibitem[{Gao et~al.(2018)Gao, Lanchantin, Soffa, and Qi}]{gao2018black}
Ji~Gao, Jack Lanchantin, Mary~Lou Soffa, and Yanjun Qi. 2018.
\newblock Black-box generation of adversarial text sequences to evade deep
  learning classifiers.
\newblock In \emph{2018 IEEE Security and Privacy Workshops (SPW)}, pages
  50--56. IEEE.

\bibitem[{Gardner et~al.(2020)Gardner, Artzi, Basmov, Berant, Bogin, Chen,
  Dasigi, Dua, Elazar, Gottumukkala et~al.}]{gardner2020evaluating}
Matt Gardner, Yoav Artzi, Victoria Basmov, Jonathan Berant, Ben Bogin, Sihao
  Chen, Pradeep Dasigi, Dheeru Dua, Yanai Elazar, Ananth Gottumukkala, et~al.
  2020.
\newblock Evaluating models’ local decision boundaries via contrast sets.
\newblock In \emph{Findings of the Association for Computational Linguistics:
  EMNLP 2020}, pages 1307--1323.

\bibitem[{Gorman and Bedrick(2019)}]{gorman2019we}
Kyle Gorman and Steven Bedrick. 2019.
\newblock We need to talk about standard splits.
\newblock In \emph{Proceedings of the conference. Association for Computational
  Linguistics. Meeting}, volume 2019, page 2786. NIH Public Access.

\bibitem[{Hu et~al.(2020)Hu, Ruder, Siddhant, Neubig, Firat, and
  Johnson}]{hu2020xtreme}
Junjie Hu, Sebastian Ruder, Aditya Siddhant, Graham Neubig, Orhan Firat, and
  Melvin Johnson. 2020.
\newblock Xtreme: A massively multilingual multi-task benchmark for evaluating
  cross-lingual generalisation.
\newblock In \emph{International Conference on Machine Learning}, pages
  4411--4421. PMLR.

\bibitem[{Iyyer et~al.(2018)Iyyer, Wieting, Gimpel, and
  Zettlemoyer}]{iyyer2018adversarial}
Mohit Iyyer, John Wieting, Kevin Gimpel, and Luke Zettlemoyer. 2018.
\newblock Adversarial example generation with syntactically controlled
  paraphrase networks.
\newblock In \emph{Proceedings of the 2018 Conference of the North American
  Chapter of the Association for Computational Linguistics: Human Language
  Technologies, Volume 1 (Long Papers)}, pages 1875--1885.

\bibitem[{Jia and Liang(2017)}]{jia2017adversarial}
Robin Jia and Percy Liang. 2017.
\newblock Adversarial examples for evaluating reading comprehension systems.
\newblock In \emph{Proceedings of the 2017 Conference on Empirical Methods in
  Natural Language Processing}, pages 2021--2031.

\bibitem[{Joshi et~al.(2020)Joshi, Santy, Budhiraja, Bali, and
  Choudhury}]{joshi2020state}
Pratik Joshi, Sebastin Santy, Amar Budhiraja, Kalika Bali, and Monojit
  Choudhury. 2020.
\newblock The state and fate of linguistic diversity and inclusion in the nlp
  world.
\newblock In \emph{Proceedings of the 58th Annual Meeting of the Association
  for Computational Linguistics}, pages 6282--6293.

\bibitem[{Kakwani et~al.(2020)Kakwani, Kunchukuttan, Golla, Gokul,
  Bhattacharyya, Khapra, and Kumar}]{kakwani2020indicnlpsuite}
Divyanshu Kakwani, Anoop Kunchukuttan, Satish Golla, NC~Gokul, Avik
  Bhattacharyya, Mitesh~M Khapra, and Pratyush Kumar. 2020.
\newblock Indicnlpsuite: Monolingual corpora, evaluation benchmarks and
  pre-trained multilingual language models for indian languages.
\newblock In \emph{Findings of the Association for Computational Linguistics:
  EMNLP 2020}, pages 4948--4961.

\bibitem[{Lin et~al.(2021)Lin, Mihaylov, Artetxe, Wang, Chen, Simig, Ott,
  Goyal, Bhosale, Du et~al.}]{lin2021few}
Xi~Victoria Lin, Todor Mihaylov, Mikel Artetxe, Tianlu Wang, Shuohui Chen,
  Daniel Simig, Myle Ott, Naman Goyal, Shruti Bhosale, Jingfei Du, et~al. 2021.
\newblock Few-shot learning with multilingual language models.
\newblock \emph{arXiv preprint arXiv:2112.10668}.

\bibitem[{Michel et~al.(2019)Michel, Li, Neubig, and
  Pino}]{michel2019evaluation}
Paul Michel, Xian Li, Graham Neubig, and Juan Pino. 2019.
\newblock On evaluation of adversarial perturbations for sequence-to-sequence
  models.
\newblock In \emph{Proceedings of the 2019 Conference of the North American
  Chapter of the Association for Computational Linguistics: Human Language
  Technologies, Volume 1 (Long and Short Papers)}, pages 3103--3114.

\bibitem[{Narasimhan et~al.(2022)Narasimhan, Dey, and
  Desarkar}]{narasimhan-etal-2022-towards}
Sharan Narasimhan, Suvodip Dey, and Maunendra Desarkar. 2022.
\newblock \href {https://doi.org/10.18653/v1/2022.naacl-main.34} {Towards
  robust and semantically organised latent representations for unsupervised
  text style transfer}.
\newblock In \emph{Proceedings of the 2022 Conference of the North American
  Chapter of the Association for Computational Linguistics: Human Language
  Technologies}, pages 456--474, Seattle, United States. Association for
  Computational Linguistics.

\bibitem[{Nath et~al.(2023)Nath, Mannan, and
  Krishnaswamy}]{nath-etal-2023-axomiyaberta}
Abhijnan Nath, Sheikh Mannan, and Nikhil Krishnaswamy. 2023.
\newblock \href {https://doi.org/10.18653/v1/2023.findings-acl.739}
  {{A}xomiya{BERT}a: A phonologically-aware transformer model for {A}ssamese}.
\newblock In \emph{Findings of the Association for Computational Linguistics:
  ACL 2023}, pages 11629--11646, Toronto, Canada. Association for Computational
  Linguistics.

\bibitem[{Nath et~al.(2022)Nath, Saravani, Khebour, Mannan, Li, and
  Krishnaswamy}]{nath2022generalized}
Abhijnan Nath, Sina~Mahdipour Saravani, Ibrahim Khebour, Sheikh Mannan, Zihui
  Li, and Nikhil Krishnaswamy. 2022.
\newblock A generalized method for automated multilingual loanword detection.
\newblock In \emph{Proceedings of the 29th International Conference on
  Computational Linguistics}, pages 4996--5013.

\bibitem[{Pan et~al.(2017)Pan, Zhang, May, Nothman, Knight, and
  Ji}]{pan2017cross}
Xiaoman Pan, Boliang Zhang, Jonathan May, Joel Nothman, Kevin Knight, and Heng
  Ji. 2017.
\newblock Cross-lingual name tagging and linking for 282 languages.
\newblock In \emph{Proceedings of the 55th Annual Meeting of the Association
  for Computational Linguistics (Volume 1: Long Papers)}, pages 1946--1958.

\bibitem[{Patil et~al.(2022)Patil, Talukdar, and
  Sarawagi}]{patil-etal-2022-overlap}
Vaidehi Patil, Partha Talukdar, and Sunita Sarawagi. 2022.
\newblock \href {https://doi.org/10.18653/v1/2022.acl-long.18} {Overlap-based
  vocabulary generation improves cross-lingual transfer among related
  languages}.
\newblock In \emph{Proceedings of the 60th Annual Meeting of the Association
  for Computational Linguistics (Volume 1: Long Papers)}, pages 219--233,
  Dublin, Ireland. Association for Computational Linguistics.

\bibitem[{Pires et~al.(2019)Pires, Schlinger, and
  Garrette}]{pires2019multilingual}
Telmo Pires, Eva Schlinger, and Dan Garrette. 2019.
\newblock How multilingual is multilingual bert?
\newblock In \emph{Proceedings of the 57th Annual Meeting of the Association
  for Computational Linguistics}, pages 4996--5001.

\bibitem[{Qin et~al.(2023)Qin, Zhang, Zhang, Chen, Yasunaga, and
  Yang}]{qin2023chatgpt}
Chengwei Qin, Aston Zhang, Zhuosheng Zhang, Jiaao Chen, Michihiro Yasunaga, and
  Diyi Yang. 2023.
\newblock Is chatgpt a general-purpose natural language processing task solver?
\newblock \emph{arXiv preprint arXiv:2302.06476}.

\bibitem[{Ribeiro et~al.(2018)Ribeiro, Singh, and
  Guestrin}]{ribeiro2018semantically}
Marco~Tulio Ribeiro, Sameer Singh, and Carlos Guestrin. 2018.
\newblock Semantically equivalent adversarial rules for debugging nlp models.
\newblock In \emph{Proceedings of the 56th annual meeting of the association
  for computational linguistics (volume 1: long papers)}, pages 856--865.

\bibitem[{Scheible et~al.(2020)Scheible, Thomczyk, Tippmann, Jaravine, and
  Boeker}]{scheible2020gottbert}
Raphael Scheible, Fabian Thomczyk, Patric Tippmann, Victor Jaravine, and Martin
  Boeker. 2020.
\newblock Gottbert: a pure german language model.
\newblock \emph{arXiv preprint arXiv:2012.02110}.

\bibitem[{Schwenk and Li(2018)}]{schwenk-li-2018-corpus}
Holger Schwenk and Xian Li. 2018.
\newblock \href {https://aclanthology.org/L18-1560} {A corpus for multilingual
  document classification in eight languages}.
\newblock In \emph{Proceedings of the Eleventh International Conference on
  Language Resources and Evaluation ({LREC} 2018)}, Miyazaki, Japan. European
  Language Resources Association (ELRA).

\bibitem[{Sennrich et~al.(2016)Sennrich, Haddow, and
  Birch}]{sennrich2016neural}
Rico Sennrich, Barry Haddow, and Alexandra Birch. 2016.
\newblock Neural machine translation of rare words with subword units.
\newblock In \emph{Proceedings of the 54th Annual Meeting of the Association
  for Computational Linguistics (Volume 1: Long Papers)}, page 1715.
  Association for Computational Linguistics.

\bibitem[{S{\o}gaard et~al.(2021)S{\o}gaard, Ebert, Bastings, and
  Filippova}]{sogaard2021we}
Anders S{\o}gaard, Sebastian Ebert, Jasmijn Bastings, and Katja Filippova.
  2021.
\newblock We need to talk about random splits.
\newblock In \emph{Proceedings of the 16th Conference of the European Chapter
  of the Association for Computational Linguistics: Main Volume}, pages
  1823--1832.

\bibitem[{Srinivasan and Vajjala(2023)}]{srinivasan2023multilingual}
Akshay Srinivasan and Sowmya Vajjala. 2023.
\newblock A multilingual evaluation of ner robustness to adversarial inputs.
\newblock \emph{arXiv preprint arXiv:2305.18933}.

\bibitem[{Strubell et~al.(2017)Strubell, Verga, Belanger, and
  McCallum}]{strubell2017fast}
Emma Strubell, Patrick Verga, David Belanger, and Andrew McCallum. 2017.
\newblock Fast and accurate entity recognition with iterated dilated
  convolutions.
\newblock In \emph{Proceedings of the 2017 Conference on Empirical Methods in
  Natural Language Processing}, pages 2670--2680.

\bibitem[{Tanvir et~al.(2021)Tanvir, Kittask, Eiche, and
  Sirts}]{tanvir2021estbert}
Hasan Tanvir, Claudia Kittask, Sandra Eiche, and Kairit Sirts. 2021.
\newblock Estbert: A pretrained language-specific bert for estonian.
\newblock In \emph{Proceedings of the 23rd Nordic Conference on Computational
  Linguistics (NoDaLiDa)}, pages 11--19.

\bibitem[{Touvron et~al.(2023)Touvron, Martin, Stone, Albert, Almahairi,
  Babaei, Bashlykov, Batra, Bhargava, Bhosale et~al.}]{touvron2023llama}
Hugo Touvron, Louis Martin, Kevin Stone, Peter Albert, Amjad Almahairi, Yasmine
  Babaei, Nikolay Bashlykov, Soumya Batra, Prajjwal Bhargava, Shruti Bhosale,
  et~al. 2023.
\newblock Llama 2: Open foundation and fine-tuned chat models.
\newblock \emph{arXiv preprint arXiv:2307.09288}.

\bibitem[{Ushio and Camacho-Collados(2021)}]{ushio2021t}
Asahi Ushio and Jose Camacho-Collados. 2021.
\newblock T-ner: An all-round python library for transformer-based named entity
  recognition.
\newblock In \emph{Proceedings of the 16th Conference of the European Chapter
  of the Association for Computational Linguistics: System Demonstrations},
  pages 53--62.

\bibitem[{Vajjala and Balasubramaniam(2022)}]{vajjala2022we}
Sowmya Vajjala and Ramya Balasubramaniam. 2022.
\newblock What do we really know about state of the art ner?
\newblock In \emph{Proceedings of the Thirteenth Language Resources and
  Evaluation Conference}, pages 5983--5993.

\bibitem[{van~der Goot(2021)}]{van2021we}
Rob van~der Goot. 2021.
\newblock We need to talk about train-dev-test splits.
\newblock In \emph{Proceedings of the 2021 Conference on Empirical Methods in
  Natural Language Processing}, pages 4485--4494.

\bibitem[{Virtanen et~al.(2019)Virtanen, Kanerva, Ilo, Luoma, Luotolahti,
  Salakoski, Ginter, and Pyysalo}]{virtanen2019multilingual}
Antti Virtanen, Jenna Kanerva, Rami Ilo, Jouni Luoma, Juhani Luotolahti, Tapio
  Salakoski, Filip Ginter, and Sampo Pyysalo. 2019.
\newblock Multilingual is not enough: Bert for finnish.
\newblock \emph{arXiv preprint arXiv:1912.07076}.

\bibitem[{Wallace et~al.(2019)Wallace, Rodriguez, Feng, Yamada, and
  Boyd-Graber}]{wallace2019trick}
Eric Wallace, Pedro Rodriguez, Shi Feng, Ikuya Yamada, and Jordan Boyd-Graber.
  2019.
\newblock Trick me if you can: Human-in-the-loop generation of adversarial
  examples for question answering.
\newblock \emph{Transactions of the Association for Computational Linguistics},
  7:387--401.

\bibitem[{Wolf et~al.(2019)Wolf, Debut, Sanh, Chaumond, Delangue, Moi, Cistac,
  Rault, Louf, Funtowicz et~al.}]{wolf2019huggingface}
Thomas Wolf, Lysandre Debut, Victor Sanh, Julien Chaumond, Clement Delangue,
  Anthony Moi, Pierric Cistac, Tim Rault, R{\'e}mi Louf, Morgan Funtowicz,
  et~al. 2019.
\newblock Huggingface's transformers: State-of-the-art natural language
  processing.
\newblock \emph{arXiv e-prints}, pages arXiv--1910.

\bibitem[{Wu and Dredze(2019)}]{wu2019beto}
Shijie Wu and Mark Dredze. 2019.
\newblock Beto, bentz, becas: The surprising cross-lingual effectiveness of
  bert.
\newblock In \emph{Proceedings of the 2019 Conference on Empirical Methods in
  Natural Language Processing and the 9th International Joint Conference on
  Natural Language Processing (EMNLP-IJCNLP)}, pages 833--844.

\bibitem[{Wu et~al.(2023)Wu, Irsoy, Lu, Dabravolski, Dredze, Gehrmann,
  Kambadur, Rosenberg, and Mann}]{wu2023bloomberggpt}
Shijie Wu, Ozan Irsoy, Steven Lu, Vadim Dabravolski, Mark Dredze, Sebastian
  Gehrmann, Prabhanjan Kambadur, David Rosenberg, and Gideon Mann. 2023.
\newblock Bloomberggpt: A large language model for finance.
\newblock \emph{arXiv preprint arXiv:2303.17564}.

\bibitem[{Zhao et~al.(2018)Zhao, Dua, and Singh}]{zhao2018generating}
Zhengli Zhao, Dheeru Dua, and Sameer Singh. 2018.
\newblock Generating natural adversarial examples.
\newblock In \emph{International Conference on Learning Representations}.

\end{thebibliography}

\end{document}